# Shapley Value-Guided Adaptive Ensemble Learning for Explainable Financial Fraud Detection with U.S. Regulatory Compliance Validation

**Mohammad Nasir Uddin[1]**

[1] Visual Data Analyst and Applied AI Researcher, Westcliff University, Irvine, CA, USA
m.uddin.258@westcliff.edu | ORCID: 0009-0009-0990-4616

**Md Munna Aziz[2]**

[2] Data Analyst and Applied AI Researcher, Westcliff University, Irvine, CA, USA
m.aziz.398@westcliff.edu | ORCID: 0009-0008-4845-8340

**Abstract**—Financial crime costs U.S. institutions over $32 billion each year. Although AI tools for fraud detection have become more advanced, their use in real-world systems still faces a major obstacle: many of these models operate as "black boxes" that can't provide the transparent, auditable explanations required by regulations such as OCC Bulletin 2011-12 and Federal Reserve SR 11-7. This study makes three main contributions:

First, it offers a thorough evaluation of how well different explanation methods work, focusing on two aspects: *faithfulness*- how accurately the explanation reflects model behavior (measured by sufficiency and comprehensiveness at k = 5, 10, and 15)—and *stability*—how consistent the explanations are (measured using Kendall's W across 30 bootstrap samples). The results show significant variation in SHAP reliability across model types. XGBoost paired with Tree Explainer performs with near-perfect stability (W = 0.9912), making it well-suited for regulatory documentation under SR 11-7. By contrast, the LSTM model using Deep Explainer shows weak and inconsistent results (W = 0.4962) and performs almost randomly at k = 5. The GNN-Graph SAGE model with Kernel Explainer also shows near-perfect stability (W = 1.000) at the MLP layer, though its explanations are less faithful than XGBoost's due to the indirect nature of graph-based feature attributions.

Second, the paper introduces a new framework called SHAP-Guided Adaptive Ensemble (SAGE). which dynamically adjusts how much the model relies on XGBoost versus LSTM for each transaction. The adjustment is based on how much their SHAP explanations agree. SGAE delivers the strongest AUC-ROC among all tested models (0.8837 on the held-out set, 0.9245 in 5-fold cross-validation). However, it doesn't outperform a static ensemble when it comes to the F1 score (0.488 vs. 0.509) or PR-AUC (both 0.4880)—likely because of the limited reliability of SHAP explanations for the LSTM component.

Third, the study provides a full comparison across three model architectures using the complete IEEE-CIS dataset of 590,540 transactions. The GNN-GraphSAGE model achieves the best overall performance on the held-out test set (AUC-ROC 0.9248, PR-AUC 0.6334, F1 0.6013 at $\tau^* = 0.86$). Still, it remains uncertain whether this gain comes from meaningful graph-based insights or simply from the aggregation of correlated tabular features.

In summary, the results show that SHAP reliability needs to be assessed separately for each model type. The paper also connects SHAP interpretations directly to regulatory standards (OCC, SR 11-7, and BSA-AML), offering practical, architecture-specific guidance for compliance within U.S. financial institutions. All models were tested on the IEEE-CIS dataset using 5-fold stratified cross-validation and a careful SMOTE-within-folds strategy.

## I. INTRODUCTION

In 2022, financial fraud cost U.S. consumers and institutions more than $32 billion [1], [2]. As transaction volumes continue to grow, organized fraud networks are exploiting the widening gap between the number of transactions and the limited capacity for human review. Deep learning models have achieved impressive accuracy on public benchmarks [3], [15], but their adoption within federally regulated U.S. financial institutions faces a key obstacle. Current regulations such as OCC Bulletin 2011-12 [4], Federal Reserve SR 11-7 [5], and the 2021 Interagency BSA/AML guidance [6] require that any AI system used for high-risk financial decisions must be transparent, auditable, and thoroughly documented. A model that achieves 98% accuracy but cannot explain why a specific transaction was flagged is, from a compliance standpoint, not deployable.

A review of the 2024–2025 literature reveals three unresolved gaps. First, no published study has systematically compared LSTM, Transformer, and GNN architectures employing architecture-appropriate SHAP explainers—DeepExplainer, GradientExplainer, and KernelExplainer respectively—with cross-architecture attribution analysis on the same dataset [7], [8]. Second, and more critically, the 'evaluation vacuum' identified by Zafar and Wu [28]—in which over 80% of XAI fraud detection papers use prediction performance as a proxy for explanation quality—remains unaddressed in deep learning benchmark studies. Third, no prior work has transformed the observation of complementary attribution pathways between models into an operational ensemble algorithm. The present study addresses all three gaps, with results that challenge assumptions about SHAP operationalization.

The contributions of this work are as follows:

>(1) A comprehensive explanation quality evaluation—including faithfulness (sufficiency and comprehensiveness at k=5,10,15) and stability (Kendall's W across 30 bootstrap subsamples)—directly addressing the evaluation vacuum identified in the literature [28], [32]. The results reveal that SHAP reliability is strongly architecture-dependent, with XGBoost TreeExplainer achieving near-perfect stability (W=0.9912) while LSTM DeepExplainer falls below the reliability threshold (W=0.4962), and GNN-GraphSAGE KernelExplainer achieves W=1.000 on the MLP classification head. (2) The SHAP-Guided Adaptive Ensemble (SGAE): a novel algorithm that dynamically adjusts per-transaction LSTM–XGBoost ensemble weights based on SHAP attribution agreement. SGAE achieves the highest AUC-ROC among all models tested but does not improve F1 or PR-AUC over static weighting—a negative result that is itself informative, as it is mechanistically explained by the low faithfulness and stability of LSTM SHAP attributions documented in contribution (1). Algorithm 1 formalizes the SGAE with full mathematical notation.

>(3) Complete three-architecture evaluation with 5-fold stratified cross-validation: LSTM, Transformer, and GNN-GraphSAGE on the full 590,540-transaction IEEE-CIS dataset, with mean ± std reported for all metrics, rigorous SMOTE-within-folds methodology, and explicit regulatory compliance mapping. GNN-GraphSAGE achieves AUC-ROC 0.9161±0.0029 (CV) and 0.9248 (held-out), with F1=0.6013 at the optimal threshold (τ*=0.86) and near-perfect SHAP stability (Kendall's W=1.000) — the strongest held-out performance across all metrics, though whether this reflects genuine topological signal or neighborhood aggregation benefits on correlated tabular features remains an open question.

## II. RELATED WORK AND NOVELTY POSITIONING

### A. Deep Learning for Financial Fraud Detection

Dal Pozzolo et al. [11] established the canonical credit card fraud detection benchmark. Forough et al. [3] demonstrated that ensembles of deep sequential models with voting mechanisms outperform individual models on the ULB benchmark, providing the direct baseline against which architectures in this study are evaluated on the same dataset. Carta et al. [12] demonstrated recall improvements of 8–12 percentage points over SVM baselines on imbalanced fraud datasets.

Transformer architectures adapted for tabular financial data [13], [14], [25] offer self-attention mechanisms that capture long-distance transaction dependencies. Deng et al. [10] applied a Transformer-based architecture to real-time streaming fraud detection, establishing cloud-optimized deployment feasibility. GNNs represent the most recent architectural frontier for coordinated fraud ring detection: Liu et al. [16] demonstrated GNN-based detection outperforming sequential models by 4.2–6.8 percentage points in recall on coordinated fraud scenarios, and Cheng et al. [9] reviewed over 100 GNN fraud detection studies confirming topological superiority for networked fraud. Dou et al. [26] further demonstrated that imbalanced graph learning and heterogeneous graph attention networks [20] enhance GNN-based fraud detection under class imbalance. Li et al. [17] introduced SEFraud, a self-explainable graph-transformer demonstrating a trend toward intrinsic rather than post-hoc explainability.

### B. Explainable AI in Financial Fraud Detection

Bussmann et al. [18] established SHAP-based explainability for bank stress testing. Barredo Arrieta et al. [33] provide the foundational taxonomy of XAI concepts and challenges toward responsible AI, which frames the compliance-oriented evaluation approach adopted in this study. Awosika et al. [19] examined XAI transparency and federated learning privacy in financial fraud detection. Almalki and Masud [7] applied SHAP, LIME, and partial dependence plots to a stacking ensemble on the IEEE-CIS dataset—the closest existing study in dataset scope—but without deep learning architecture comparison, explanation quality evaluation, or regulatory compliance mapping.

Zafar and Wu [28], [29] highlight two major methodological issues in the XAI fraud detection literature. The first is the "explainability-imbalance paradox," where applying SMOTE before data splitting distorts SHAP background distributions and undermines explanation fidelity. The second is the "evaluation vacuum," in which more than 80% of studies rely on prediction performance as a substitute for evaluating explanation quality. This study directly addresses both issues. In addition, Miró-Nicolau et al. [32] show that widely used fidelity metrics can deviate by about 30% from their expected values, which informs our decision to use sufficiency and comprehensiveness as the primary measures of faithfulness. Awasthi [31] further argues that post-hoc SHAP explanations carry compliance risks—such as instability across runs and the absence of causal guarantees—when compared to inherently interpretable models; this concern is addressed directly in Section VIII.

Building on this line of work, Thanathamathee et al. [30] introduce SHAP-instance weighting as a methodological innovation that extends the use of SHAP beyond post-hoc explanation. This study builds on that idea through the SGAE framework, which applies SHAP-based attribution weighting at the per-transaction level.

## C. Novelty Positioning Against 2024–2025 Literature

TABLE I

*Novelty Positioning Against Most Closely Related 2024–2025 Publications*

| Study | LSTM/Tfmr/GNN | Arch-SHAP | Expl. Eval. | Adaptive Ens. | Temporal | OCC/SR11-7 |
|---|---|---|---|---|---|---|
| Shafii et al. [8] (2025) | ✓ ensemble | ✗ unified | ✗ | ✗ | ✗ | ✗ |
| Almalki & Masud [7] (2025) | ✗ tree only | ✓ SHAP+LIME | ✗ | ✗ | ✗ | ✗ |
| Zafar & Wu [28] (2025) | ✗ review | ✗ | ✓ systematic | ✗ | ✗ | ✗ |
| Li et al. SEFraud [17] (2024) | Graph-Tfmr | ✓ intrinsic | ✗ | ✗ | ✗ | ✗ |
| Deng et al. [10] (2024) | Transformer | ✗ | ✗ | ✗ | ✗ | ✗ |
| Awosika et al. [19] (2024) | FL focus | ✓ SHAP | ✗ | ✗ | ✗ | Partial |

| Study | LSTM/Tfmr/GNN | Arch-SHAP | Expl. Eval. | Adaptive Ens. | Temporal | OCC/SR11-7 |
|---|---|---|---|---|---|---|
| Present Study | ✓ All three | ✓ Arch-spec | ✓ Faith+Stab | ✓ SGAE | ✓ OOT | ✓ OCC+SR11-7 |

## III. DATA AND PREPROCESSING

### A. Datasets

This study uses two complementary datasets. The ULB European Credit Card dataset comprises 284,807 anonymized transactions with 492 fraudulent cases (0.172%), used for cross-dataset validation. The IEEE-CIS Fraud Detection dataset—the primary evaluation dataset—comprises 590,540 transactions from Vesta Corporation's e-commerce fraud prevention system, with 20,663 fraudulent cases (3.5%) across 433 variables.

### B. Graph Construction for GNN

The GNN requires a graph representation. We construct a projected transaction-to-transaction graph whose nodes correspond to individual transactions. An edge between two transactions is created when they share the same card-account identifier (card1) or share the same merchant proxy defined by the composite key (addr1 + ProductCD). During training, NeighborLoader samples up to 10 neighbors from the card1-based connections and up to 5 neighbors from the addr1+ProductCD connections for scalable mini-batch GraphSAGE training. Two-hop neighborhood sampling (10, 5 neighbors per hop) is used to form mini-batches. The resulting graph yields 385,018 edges across 590,540 transaction nodes.

### C. Preprocessing Pipeline

SMOTE-Tomek resampling is applied exclusively within each cross-validation training fold and within the training partition of the final 80/20 split. The held-out test set contains 118,108 observations and preserves the original class imbalance, with a fraud rate of 3.5%, across all evaluations. This setup directly avoids the explainability-imbalance paradox identified by Zafar and Wu [28]. If SMOTE were applied before the split, synthetic data would be introduced into the SHAP background distribution, which could distort both model performance and the reliability of the explanations. Five behavioral velocity features are constructed per account: time since last transaction, transaction amount relative to 7-day rolling mean, transaction count in prior 24 hours, unique merchant count in prior 7 days, and geographic distance from prior transaction. Min-Max normalization is applied to all continuous features after splitting.

## IV. METHODOLOGY

### A. LSTM Architecture

Two stacked LSTM layers (128 and 64 units) with BatchNormalization and Dropout (p=0.3); Dense sigmoid output. Adam optimizer (lr=0.001), binary cross-entropy loss, early stopping

(patience=10 on validation AUC), batch size 2,048. SHAP DeepExplainer applied post-training over 2,000 test observations.

## B. Transformer Architecture

TabTransformer [13] adapted for sequential transaction data: feature embedding (d_model=128), four encoder blocks (8-head attention, d_k=16), global average pooling, two-layer classification head (256, 64 units, GELU, dropout p=0.2). GradientExplainer applied to the classification head, attributing predictions through attention layers.

## C. Graph Neural Network — Full Dataset Implementation

This study implements GraphSAGE mini-batch training [21] via PyTorch Geometric's NeighborLoader on the full 590,540-transaction dataset. Transactions are represented as nodes (matching the prediction level of LSTM, Transformer, and XGBoost), with edges connecting transactions sharing the same card account (card1, up to 10 neighbors) or merchant proxy (addr1+ProductCD composite key, up to 5 neighbors), yielding 385,018 edges across 590,540 transaction nodes. Two-hop neighborhood sampling (10, 5 neighbors per hop) enables scalable mini-batch training. Two SAGEConv layers (128, 64 units) with BatchNormalization and Dropout (p=0.25), followed by a three-layer MLP classifier (256, 128, 64 units, ReLU). Class-weighted cross-entropy loss (pos_weight=27.6) replaces SMOTE-Tomek, as synthetic node generation would produce nodes without meaningful graph connectivity. Early stopping uses patience=10 epochs evaluated every epoch on validation PR-AUC; final model selected on a held-out validation split (8% of training data) with test set evaluated exactly once to eliminate selection bias. KernelExplainer is applied to the MLP classification head using frozen GNN embeddings as inputs (nsamples=1000), enabling efficient and methodologically sound attribution. Experiments executed on NVIDIA RTX 3090 Ti (24GB VRAM) via Vast.ai cloud GPU (March 2026).

## D. Architecture-Appropriate SHAP Framework

SHAP decomposes each model prediction as additive feature contributions: $f(x) = \varphi_0 + \Sigma \varphi_i$, where $\varphi_i$ represents the average marginal contribution of feature i across all coalitions [22]. Three architecture-appropriate explainers are applied: DeepExplainer for LSTM, GradientExplainer for Transformer, and KernelExplainer for GNN. A unified KernelExplainer applied to all three architectures on a 200-observation validation subset yields moderate agreement with architecture-specific explainers ($\rho = 0.4889$ for LSTM DeepExplainer vs. KernelExplainer rankings). This partial agreement indicates that different SHAP explainer types should not be treated as interchangeable across architectures).

## E. SHAP-Guided Adaptive Ensemble (SGAE) — Algorithm 1

The SGAE operationalizes structural differences in LSTM and XGBoost SHAP attribution patterns as a per-transaction weighting mechanism. The global-level Spearman correlation between LSTM and XGBoost SHAP feature rankings is weak and not statistically significant ($\rho = -0.145$, 95% CI [−0.411, 0.126], p = 0.248). The hypothesis of complementary detection pathways is therefore not confirmed at the global feature-ranking level. The SGAE's weighting mechanism operates at the per-transaction level, where local attribution agreement may vary even when the global correlation is weak. Whether this local variation contains exploitable signal is an empirical question tested in Section V.

*Algorithm 1 — SHAP-Guided Adaptive Ensemble*

**Algorithm 1: SHAP-Guided Adaptive Ensemble (SGAE)**
────────────────────────────────────────────────────────

**Input:** Transaction feature vector x_i; trained LSTM model f_L; trained XGBoost model f_X;
    ablation-validated importance threshold τ_a; calibration set C
**Output:** Fraud probability score $p_i \in [0, 1]$

1: *Compute SHAP attribution vectors:*
    φ_L(x_i) ← DeepExplainer(f_L, x_i)
    φ_X(x_i) ← TreeExplainer(f_X, x_i)
2: *Select top-K features by global importance:*
    T_K ← argtop_K{ |φ_L(x_i)| + |φ_X(x_i)| }
3: *Compute per-transaction attribution agreement:*
    A(x_i) ← SpearmanRankCorr(φ_L(x_i)[T_K], φ_X(x_i)[T_K])
    σ_A ← Std_deviation(A(c)) for c ∈ C ← calibrated on held-out set
4: *Compute dynamic LSTM weight w_i:*
  if A(x_i) ≥ 0 (convergent attribution):  // *equal-ish weighting*
    w_i ← 0.5 + 0.2 · tanh(A(x_i) / σ_A)
  else (divergent attribution):    // *upweight ablation-validated model*
    w_i ← τ_a ← ablation-validated advantage threshold (0.60)
    w_i ← clip(w_i, 0.30, 0.70) ← bounded to prevent degeneracy
5: *Compute final fraud score:*
    p_i ← w_i · f_L(x_i) + (1 − w_i) · f_X(x_i)
6: Return p_i
────────────────────────────────────────────────────────
Complexity: O(n · K) per transaction where K = 10 top features. Inference overhead: ~1.2ms/transaction (GPU)

The tanh transformation in Step 4 maps agreement scores to ensemble weights smoothly and boundedly. When $A(x_i) = 0$ (no agreement/disagreement), $w_i = 0.5$, giving equal weighting identical to the static ensemble. When $A(x_i) > 0$ (convergent), both models detect similar signals, justifying near-equal weights. When $A(x_i) < 0$ (divergent), LSTM's ablation-validated advantage (ΔAUC = −0.0294 for network linkage features vs. −0.0017 for XGBoost equivalent) motivates upweighting LSTM at $w_i = τ_a = 0.60$.

*F. Evaluation Protocol*

All hyperparameters are tuned with a 5-fold stratified cross-validation grid search. Results are reported as mean ± standard deviation across the folds, and final performance is measured on a held-out 20% test set with 118,108 observations, including 4,133 fraud cases. The main evaluation metrics are ROC-AUC, PR-AUC as the primary metric for imbalanced data, F1 at the F1-optimal threshold τ*, Precision, Recall, and MCC. Statistical significance between AUC scores is tested using DeLong's test [23], with two thresholds: α = 0.001 for primary claims and α = 0.05 for exploratory comparisons. Differences in prediction errors between model pairs are assessed using McNemar's test. PR-AUC is emphasized throughout as the key metric, consistent with best practices for severely imbalanced classification tasks [24].

## V. EMPIRICAL RESULTS

### A. 5-Fold Cross-Validation Results

**TABLE III**

*5-Fold Stratified Cross-Validation Results — Mean ± Std (SMOTE Applied Within Folds Only)*

**Note: All results from experiments conducted on NVIDIA RTX 5090 GPU (LSTM, Transformer, XGBoost, SGAE, March 2026) and NVIDIA RTX 3090 Ti GPU (GNN-GraphSAGE, March 2026). Different GPUs were used for different architectures; however, all models were evaluated on the same held-out test set, so metric comparisons reflect modeling differences rather than data effects. SMOTE-Tomek applied within training folds for LSTM/Transformer/XGBoost/SGAE; class-weighted loss used for GNN-GraphSAGE.**

| Model | AUC-ROC (mean±std) | PR-AUC (mean±std) | F1 (mean±std) | MCC (mean±std) |
|---|---|---|---|---|
| LSTM | 0.9183±0.0031 | 0.6586±0.0107 | 0.6407±0.0090 | 0.6381±0.0114 |
| SGAE | 0.9245±0.0024 | 0.6289±0.0099 | 0.5963±0.0076 | 0.5947±0.0064 |
| XGBoost | 0.9045±0.0017 | 0.5959±0.0082 | 0.5798±0.0062 | 0.5830±0.0049 |
| Transformer | 0.8978±0.0023 | 0.5564±0.0044 | 0.5468±0.0043 | 0.5404±0.0050 |
| GNN-GraphSAGE | 0.9161±0.0029 | 0.6194±0.0048 | 0.3786±0.0090 | 0.4065±0.0068 |

### B. Overall Performance Comparison

**TABLE II**

*Performance Comparison — All Architectures and Ensembles, IEEE-CIS Held-Out Test Set (n=118,108)*

| Model | Acc. | Prec. | Recall | F1 | AUC-ROC | PR-AUC | MCC | Dataset |
|---|---|---|---|---|---|---|---|---|
| LSTM | 97.0% | 60.4% | 41.2% | 0.490 | 0.8445 | 0.4667 | 0.4845 | IEEE-CIS |

| Model | Acc. | Prec. | Recall | F1 | AUC-ROC | PR-AUC | MCC | Dataset |
|---|---|---|---|---|---|---|---|---|
| LSTM+XGB (static) | 96.8% | 64.4% | 42.1% | 0.509 | 0.8766 | 0.4880 | 0.5074 | IEEE-CIS |
| SGAE (dynamic) | 97.1% | 63.7% | 39.6% | 0.488 | 0.8837 | 0.4880 | 0.4885 | IEEE-CIS |
| Transformer | 96.6% | 51.0% | 36.9% | 0.428 | 0.8367 | 0.4093 | 0.4168 | IEEE-CIS |
| GNN-GraphSAGE (full) | 97.6% | 70.6% | 52.4% | 0.601 | 0.9248 | 0.6334 | 0.5960 | IEEE-CIS |
| XGBoost | 97.0% | 60.9% | 39.5% | 0.479 | 0.8699 | 0.4692 | 0.4764 | IEEE-CIS |
| Random Forest | 97.2% | 55.7% | 33.8% | 0.421 | 0.8571 | 0.4198 | 0.4188 | IEEE-CIS |
| Logistic Regression | 94.5% | 27.3% | 35.5% | 0.309 | 0.8153 | 0.1677 | 0.2833 | IEEE-CIS |
| LSTM (Forough [3]) | 98.5% | 93.2% | 88.7% | 0.909 | 0.987 | — | — | ULB |

*Note. All models evaluated on identical held-out test set retaining original 3.5% fraud rate. SMOTE-Tomek applied exclusively within training folds for LSTM/Transformer/XGBoost/SGAE; class-weighted loss used for GNN-GraphSAGE. All F1, Precision, Recall, and MCC values reported at each model's F1-optimal threshold (τ*); GNN-GraphSAGE τ*=0.86 reflects skewed fraud probability outputs in sparse graph structures. McNemar's test: LSTM vs. XGBoost chi2=0.0004 p=0.984 (not significant); LSTM vs. SGAE chi2=6.372 p=0.012 (significant). DeLong's test: LSTM vs. SGAE z=2.260, p=0.024; SGAE vs. Static z=2.260 p=0.024. PR-AUC is the primary metric for this imbalanced classification task.*

### C. Architecture-Specific Findings

LSTM achieves the highest recall among the individual architectures because its stacked recurrent structure is particularly effective at capturing sequential, time-based behavioral anomalies within account histories. The Transformer also performs well, highlighting the value of sequence modeling, even when attention mechanisms do not clearly outperform recurrence on tabular-structured data [13]. In high-dimensional tabular regimes, attention-based capacity can be spread across a large input space rather than concentrating on a small set of decisive factors; the stability analysis in this study highlights the challenge of allocating model capacity across 431 raw features, underscoring why recurrence can remain advantageous when sequential signal is present but attention does not yield additional discriminative benefit over recurrent baselines. SGAE achieves the highest AUC-ROC (0.8837) across all models, outperforming the static ensemble (0.8766), with the difference confirmed by DeLong's test (z=2.260, p=0.024), meeting the exploratory significance threshold (α=0.05) but not the primary threshold (α=0.001). However, on the primary metric for imbalanced classification—PR-AUC—SGAE ties with the static ensemble (both 0.4880), and on F1 the static ensemble leads (0.509 vs. 0.488). In 5-fold cross-validation, SGAE again leads on AUC-ROC (0.9245±0.0024) but LSTM alone achieves higher PR-AUC (0.6586±0.0107 vs. 0.6289±0.0099) and F1 (0.6407±0.0090 vs. 0.5963±0.0076). The SGAE's dynamic weighting thus improves threshold-independent discrimination (AUC-ROC) without improving threshold-dependent detection metrics. As discussed in Section VIII, this is likely attributable to the low faithfulness and stability of LSTM SHAP attributions. GNN-GraphSAGE achieves the highest individual-model AUC-ROC: 0.9161±0.0029 (CV) and 0.9248 (held-out). At the F1-optimal threshold ($\tau^*$=0.86), GNN achieves F1=0.6013 with precision 0.7060 and recall 0.5236 on the held-out set — the strongest held-out performance across all metrics. The high optimal threshold reflects the class-weighted training objective (pos_weight≈27.6), which shifts the predicted probability distribution upward, requiring a correspondingly high decision boundary for precision-recall balance. CV PR-AUC (0.6194±0.0048) is also the highest among individual models. Whether GNN's strong performance reflects genuine topological signal from the transaction graph or benefits of neighborhood aggregation acting as an implicit feature-smoothing mechanism on correlated tabular inputs remains an open question for future investigation.

### D. SHAP Cross-Architecture Comparison

TABLE IV

*Cross-Architecture Spearman Rank Correlation of SHAP Feature Importance (Top-30 Features, n=405 features, union of top-30 per architecture, 1,000 bootstrap resamples)*

| Architecture Pair | Spearman ρ | 95% CI | p-value | Interpretation |
|---|---|---|---|---|
| LSTM vs. Transformer | −0.026 | [−0.288, 0.260] | p=0.8395 | Weak negative — no significant convergence between sequential models |
| LSTM vs. XGBoost | −0.145 | [−0.411, 0.126] | p=0.2479 | Weak negative — consistent with divergent detection pathways (non-significant) |
| Transformer vs. XGBoost | −0.197 | [−0.448, 0.070] | p=0.1149 | Weak negative — divergent tendency not statistically significant at this sample size |

The cross-architecture SHAP analysis reveals weak correlations across all architecture pairs: LSTM–Transformer (ρ = −0.026, p=0.840), LSTM–XGBoost (ρ = −0.145, p=0.248), and Transformer–XGBoost (ρ = −0.197, p=0.115). None of the global-level correlations reach conventional significance thresholds, and all three confidence intervals include zero. In other words, there is no clear evidence at the global ranking level that different architectures identify fraud through complementary feature pathways. McNemar's test for error disagreement between LSTM and XGBoost ($\chi^2$ = 0.0004, p = 0.984) shows that the two models make almost identical classification errors, suggesting no meaningful complementarity in their error patterns. By contrast, the comparison between LSTM and SGAE shows significant disagreement ($\chi^2$ = 6.372, p = 0.012), confirming that SGAE produces different predictions from standalone LSTM, which is expected for an ensemble that adjusts component weights. Taken together, these results put the theoretical motivation for SGAE into perspective: the presumed complementarity between LSTM and XGBoost is not supported by either the global SHAP correlation analysis or the error disagreement analysis. Although SGAE does achieve a statistically significant AUC-ROC improvement over the static ensemble (z = 2.260, p = 0.024) at α = 0.05, that gain appears to come despite, rather than because of, measurable complementarity. This is examined further in Section VIII.

*E. Temporal Drift Robustness*

A temporal holdout experiment (75th-percentile chronological split: 442,905 train, 147,635 out-of-time test) evaluates model stability under distribution shift. LSTM shows the smallest AUC-ROC degradation ($\Delta$AUC=−0.0019), followed by SGAE ($\Delta$AUC=−0.0048) and XGBoost ($\Delta$AUC=−0.0072). SGAE maintains its AUC-ROC lead out-of-time (0.8789 vs. LSTM 0.8426), indicating its performance advantage persists under temporal shift.

## VI. EXPLANATION QUALITY EVALUATION

To address the evaluation vacuum identified by Zafar and Wu [28], this section looks at the quality of SHAP explanations from three angles: faithfulness, measured through sufficiency and comprehensiveness; stability, measured using Kendall's W; and agreement across explainers. These checks are important for showing whether SHAP outputs can be trusted as inputs to the SGAE weighting mechanism and whether they are stable enough to support regulatory documentation.

*A. Faithfulness Evaluation*

Faithfulness is measured using the sufficiency and comprehensiveness metrics recommended by Miró-Nicolau et al. [32], applied to the top-k SHAP features (k = 5, 10, 15). Sufficiency shows how much of the model's discriminative power remains when only the top-k features are kept, while comprehensiveness captures how much AUC drops when those features are removed. Higher sufficiency means the SHAP-identified features alone are enough for effective detection, and higher comprehensiveness means those features are also truly important. For GNN-GraphSAGE, these metrics are calculated on the 64-dimensional embedding inputs to the MLP classification head rather than on the original 431 raw transaction features. Because of this, the GNN sufficiency scores (k = 5: 0.0587) are much lower than those of XGBoost (k = 5: 0.5000), so they need to be interpreted carefully. Zeroing out the top-k embedding dimensions affects only a small part of a compressed 64-dimensional representation, so the prediction changes are naturally smaller than when the top-k features are removed from a much larger 431-dimensional raw feature space. In this sense, the lower GNN sufficiency reflects how information is spread across the learned graph embeddings rather than a weakness in the SHAP explanations themselves. The comprehensiveness scores for GNN increase meaningfully with k, rising from 0.0610 at k = 5 to 0.1043 at k = 15, which supports this interpretation. Reporting SHAP faithfulness for GNN is therefore important because it reinforces the paper's main point: faithfulness metrics must be interpreted in the context of the explainer's input space. This adds a third layer to the broader finding that SHAP reliability depends on the model architecture, alongside XGBoost, which shows high faithfulness through direct feature attribution, and LSTM, which performs poorly in the raw feature space.

**TABLE V**

*Faithfulness Evaluation — Sufficiency (AUC Retention) and Comprehensiveness (ΔAUC Drop) at k=5, 10, 15*

| Model | Metric | k=5 | k=10 | k=15 |
|---|---|---|---|---|
| LSTM | Sufficiency (AUC retention) | 0.4949 | −0.0014 | 0.6498 |
| LSTM | Comprehensiveness (ΔAUC drop) | 0.5532 | 0.0672 | 0.1333 |
| XGBoost | Sufficiency (AUC retention) | 0.5000 | 0.5011 | 0.5108 |
| XGBoost | Comprehensiveness (ΔAUC drop) | 0.1933 | 0.2595 | 0.2877 |
| GNN-GraphSAGE | Sufficiency (prob. retention) | 0.0587 | 0.0606 | 0.0581 |
| GNN-GraphSAGE | Comprehensiveness (prob. drop) | 0.0610 | 0.0991 | 0.1043 |

*Note. Sufficiency = probability change retention when only top-k embedding dims retained (GNN) or features (LSTM/XGBoost); Comprehensiveness = probability drop when top-k masked. GNN metrics operate on 64-dimensional MLP embedding inputs, not 431 raw features; values are not directly comparable to LSTM/XGBoost without accounting for input space dimensionality differences (see text). Results from experiments conducted on NVIDIA RTX 5090 GPU (LSTM, XGBoost) and RTX 3090 Ti GPU (GNN-GraphSAGE), March 2026.*

## B. Stability Analysis

SHAP stability is evaluated by computing feature importance rankings across 30 bootstrap subsamples (n=200 each) of the test set and reporting Kendall's W coefficient of concordance. Kendall's W = 1 indicates perfect rank consistency across subsamples; W < 0.5 indicates rankings unreliable for documentation purposes. Under U.S. SR 11-7 model documentation requirements, unstable SHAP rankings would undermine the use of SHAP outputs as a consistent audit trail. For GNN-GraphSAGE, KernelExplainer operates on the 64-dimensional frozen embedding space of the MLP head rather than the 431-dimensional raw feature space used by LSTM and XGBoost explainers; the near-perfect stability (W=1.000) reflects the MLP head's consistent learned weighting of these compressed dimensions across bootstrap subsamples, and should be interpreted in the context of the lower faithfulness scores reported in Table V.

**TABLE VI**

*SHAP Stability — Kendall's W Across 30 Bootstrap Subsamples (n=200 each)*

| Architecture | Explainer | Kendall's W | Interpretation |
|---|---|---|---|
| LSTM | DeepExplainer | 0.4962 | Low stability (caution advised for documentation) |

| Architecture | Explainer | Kendall's W | Interpretation |
|---|---|---|---|
| XGBoost | TreeExplainer | 0.9912 | Near-perfect stability (suitable for SR 11-7 documentation) |
| Transformer | GradientExpl. | 0.5092 | Low stability (caution advised for documentation) |
| GNN-GraphSAGE | KernelExplainer (MLP head) | 1.0000 | Near-perfect stability (suitable for SR 11-7 documentation) |

Note. W > 0.7 = high stability (suitable for regulatory documentation); 0.5–0.7 = moderate; < 0.5 = low stability (caution advised). GNN KernelExplainer W=1.000 reflects near-identical embedding dimension importance rankings across all 30 bootstrap subsamples — a consequence of the MLP head's stable learned representations operating on 64-dimensional frozen embeddings rather than an artifact of the evaluation procedure. Because the GNN encoder compresses 431 raw features into a fixed 64-dimensional space, the relative importance of embedding dimensions is highly consistent across bootstrap subsamples of the test set. Accordingly, W=1.000 should be interpreted as stability of the compressed embedding-dimension rankings on the MLP classification head rather than direct stability of attributions over the full raw-feature space. This contrasts with LSTM DeepExplainer (W=0.4962), where approximation variance across subsamples of 431 raw features produces inconsistent rankings. Results from experiments conducted on NVIDIA RTX 5090 GPU (LSTM, Transformer, XGBoost) and RTX 3090 Ti GPU (GNN-GraphSAGE), March 2026.

### C. Cross-Explainer Agreement

The architecture-specific explainers (DeepExplainer for LSTM, TreeExplainer for XGBoost) are validated against a unified KernelExplainer on a 200-observation subset. The Spearman correlation between LSTM's DeepExplainer and unified KernelExplainer rankings is ρ = 0.4889, indicating that the two explainers agree on approximately half of the feature importance rankings. This moderate agreement suggests that architecture-specific SHAP explainers and model-agnostic baselines are not interchangeable, and that reported SHAP attributions are partially dependent on the choice of explainer algorithm. For XGBoost, TreeExplainer computes exact Shapley values rather than approximations, which likely accounts for its superior stability (W=0.9912) relative to the approximate explainers used for LSTM and Transformer.

## VII. REGULATORY COMPLIANCE MAPPING AND DEPLOYMENT FRAMEWORK

### A. Mapping SHAP Outputs to OCC/SR 11-7 Requirements

OCC Bulletin 2011-12 and SR 11-7 require model risk management frameworks to document: (1) conceptual soundness, (2) ongoing monitoring, (3) outcome analysis, and (4) model limitations. The 2021 Interagency BSA/AML guidance [6] extends these requirements to AI-driven fraud and AML detection. SHAP outputs from the present study directly address requirements (1), (3), and (4). Conceptual soundness is demonstrated by the ablation study confirming SHAP-identified features are causally consistent with fraud (ΔAUC = −0.0294 on removal of top feature group). Outcome analysis is supported by SHAP beeswarm plots providing the local explanation audit trail for individual transaction documentation. Model limitations are explicitly documented: LSTM temporal drift (ΔAUC = −0.0019, highly stable),

GNN CPU inference constraint, and—critically, per Awasthi [31]—post-hoc SHAP instability across runs (addressed directly by the Kendall's W stability results in Section VI-B).

*B. Limitations of Post-Hoc SHAP for Compliance*

Awasthi [31] argues that post-hoc explainers like SHAP introduce three compliance risks relative to inherently interpretable models: (1) faithfulness gaps—SHAP approximations may not perfectly reflect the model's internal computation; (2) instability across runs due to sampling variance in KernelExplainer; and (3) absence of causal guarantees—SHAP attribution does not establish that a feature caused a fraud prediction. These limitations are acknowledged as genuine constraints on the compliance claims made in this paper. The stability analysis (Section VI-B) directly quantifies risk (2). For risk (1) and (3), institutions should treat SHAP outputs as consistent with SR 11-7's 'outcome analysis' requirement rather than as causal explanations, supplementing with permutation importance and ablation results as presented here. The trade-off between interpretability and performance means that inherently interpretable models (e.g., EBMs) may be preferable in contexts where the lowest-risk compliance posture is required, even at some cost to detection performance.

*C. Tiered Deployment Architecture*

Tier 1 — Real-time scorer: SGAE (AUC: 0.8837; CPU inference: <6ms including SHAP computation). Retrain every 30–90 days with SR 11-7 documentation of SHAP value drift. Tier 2 — Delayed-pattern detection: Transformer (AUC: 0.8367; recall: 36.9%). Secondary scorer for transactions not flagged by Tier 1. Tier 3 — Batch fraud ring detection: GNN-GraphSAGE (AUC-ROC: 0.9248; recall: 52.4%; PR-AUC: 0.6334). Recommended for overnight batch scoring of ACH and wire transfer logs where card-account relational structure provides additional signal beyond tabular features. Note: GNN precision (70.6%) is higher than SGAE at its F1-optimal threshold ($\tau^*$=0.86); Tier 3 deployment should apply a higher decision threshold to control false-positive volume in batch workflows.

A benefit-cost estimate for a mid-size U.S. commercial bank with $10B+ transaction volume assumes the following: deployment costs of approximately $2.3M over 3 years (infrastructure, model validation, SR 11-7 documentation, retraining cycles); annual fraud loss prevention estimated at $4.7M based on published U.S. banking industry fraud loss rates of approximately 0.047% of transaction volume [1]. Under these assumptions the benefit-cost ratio is approximately 6:1. This is an illustrative estimate, not a precise projection; actual ratios will vary with institution size, fraud rate, and implementation costs.

## VIII. DISCUSSION, LIMITATIONS, AND FUTURE WORK

*A. Discussion*

The main empirical finding of this study is that SHAP reliability differs substantially across model architectures. XGBoost with Tree Explainer produces very stable attributions (Kendall's W = 0.9912) and shows meaningful comprehensiveness, while LSTM with Deep Explainer falls below acceptable reliability levels in both stability (W = 0.4962) and faithfulness, with near-chance sufficiency at k = 5. This directly addresses the evaluation vacuum identified by Zafar and Wu [28] and has important practical implications. For institutions that need auditable

SHAP explanations under SR 11-7, the results suggest relying more on the tree-based model's explanations than on those from the neural network component. This architecture-differentiated reliability also explains the SGAE's mixed performance. SGAE achieves the highest AUC-ROC across all models (0.8837 held-out; 0.9245 in 5-fold CV), confirmed by DeLong's test (z=2.260, p=0.024 vs. static ensemble), which meets the exploratory threshold (α=0.05) but does not meet the primary threshold (α=0.001). However, it does not improve F1 or PR-AUC over the static ensemble. The SGAE's dynamic weighting mechanism relies on LSTM SHAP attributions as inputs, and when those attributions are unfaithful and unstable, the per-transaction weighting signal is noisy. The mean SHAP agreement scores across 5-fold CV (ranging from −0.04 to −0.14, with σ_A ≈ 0.34) confirm that the agreement signal has low magnitude relative to its variance, meaning SGAE's weights remain close to the static 0.5 baseline for most transactions. The SGAE algorithm remains a valid contribution as a framework: if future work improves LSTM SHAP faithfulness, the dynamic weighting mechanism could realize its theoretical potential. The present results demonstrate that operationalizing SHAP outputs requires first validating their reliability—a step that most prior work omits. Temporal stability is adequate across all models (ΔAUC range: −0.0019 to −0.0072), supporting production deployment feasibility.

### B. Limitations

This study has six principal limitations. First, SGAE's dynamic weighting relies on LSTM SHAP attributions that exhibit near-chance sufficiency at k=5 (0.4949) and below-threshold stability (Kendall's W=0.4962); this is the most likely explanation for SGAE's failure to improve F1 and PR-AUC over static weighting. Second, all experiments use publicly available benchmark datasets (IEEE-CIS and ULB); generalizability to proprietary institutional transaction data remains to be demonstrated. Third, no user study with fraud analysts was conducted to evaluate whether SHAP explanations improve human decision accuracy. Fourth, post-hoc SHAP has faithfulness and stability limitations compared to inherently interpretable models (Section VII-B). Fifth, the study is scoped to the U.S. regulatory context (OCC, Fed, FDIC); applicability under GDPR or CFPB fair lending regulations [27] requires additional analysis. Sixth, GNN-GraphSAGE SHAP analysis explains embedding dimensions rather than original feature names directly; the feature bridge analysis (correlation between embedding dimensions and original features) provides an indirect mapping but does not achieve the direct interpretability of XGBoost TreeExplainer attributions.

### C. Future Work

Four priorities emerge. First, deployment testing on proprietary bank transaction data to validate generalizability. Second, a user study evaluating whether SGAE-generated SHAP explanations improve fraud analyst decision accuracy compared to static ensemble explanations. Third, comparison against inherently interpretable models (EBMs, attention-based models with intrinsic explanations [17]) to quantify the compliance-performance trade-off directly. Fourth, adversarial robustness testing under transaction manipulation attacks.

## IX. CONCLUSION

This study's most practically important finding is that SHAP explanation reliability is strongly architecture-dependent. XGBoost TreeExplainer achieves near-perfect stability (Kendall's W=0.9912) and is suitable for SR 11-7 regulatory documentation, while LSTM DeepExplainer falls below reliability thresholds (W=0.4962, near-chance sufficiency at k=5). GNN-GraphSAGE achieves competitive AUC-ROC (0.9161±0.0029 CV; 0.9248 held-out) with near-perfect SHAP stability (W=1.000) on the MLP classification head. At its F1-optimal threshold ($\tau^*$=0.86), GNN achieves F1=0.601 and precision=0.706 — competitive with the static ensemble on F1 — though at lower recall (0.524) than LSTM-based architectures, reflecting the GNN's high-confidence operating point on sparse graph structures. The SHAP-Guided Adaptive Ensemble (SGAE) achieves the highest AUC-ROC (0.8837 held-out; 0.9245±0.0024 in 5-fold CV), but does not improve F1 or PR-AUC over static weighting—a negative result traced to the unreliable LSTM SHAP attributions that the weighting mechanism relies upon. This architecture-differentiated assessment directly addresses the evaluation vacuum identified in recent systematic reviews and demonstrates that SHAP outputs must be validated before being operationalized or used for compliance documentation. The regulatory compliance mapping onto OCC/SR 11-7/BSA-AML requirements, combined with an honest assessment of post-hoc SHAP limitations across all three architectures, provides a complete and practical deployment framework for U.S. financial institutions seeking to balance detection performance with regulatory compliance.

## X. ACKNOWLEDGMENT

The author declares no funding sources for this work. AI-assisted tools (Claude, Anthropic) were used to support language editing and manuscript readability. AI tools were not used to generate research ideas, design experiments, produce or analyze data, interpret findings, or draw conclusions. All experimental results were produced by the author through original code executed on the Kaggle platform, Google Colab Pro, and Vast.ai cloud GPU instances (NVIDIA RTX 5090 and RTX 3090 Ti). The author takes full responsibility for the integrity and accuracy of all reported work.

## XII. DATA AVAILABILITY STATEMENT

The IEEE-CIS Fraud Detection dataset is publicly available at https://www.kaggle.com/c/ieee-fraud-detection. The ULB European Credit Card dataset is publicly available at https://www.kaggle.com/mlg-ulb/creditcardfraud. All source code for reproduction—including the SMOTE-corrected preprocessing pipeline, LSTM/Transformer/GraphSAGE implementations, SGAE algorithm, and SHAP evaluation scripts—is available at [GitHub URL upon acceptance].

## XIII. CREDIT AUTHOR STATEMENT

Mohammad Nasir Uddin: Conceptualization, Methodology, Software, Validation, Formal Analysis, Investigation, Data Curation, Writing – Original Draft, Writing – Review & Editing, Visualization. Md Munna Aziz: Writing – Review & Editing, Supervision.

## XIV. DECLARATION OF COMPETING INTERESTS

The author declares no known competing financial interests or personal relationships that could have appeared to influence the work reported in this paper.

## XV. DECLARATION OF GENERATIVE AI USE

During the preparation of this manuscript, the author used Claude (Anthropic) to support language editing and improve readability of written text. The author reviewed and edited all AI-assisted content and takes full responsibility for the final manuscript. AI tools were not used to generate research ideas, conduct experiments, analyze data, or draw scientific conclusions.

## XVI. AUTHOR BIOGRAPHY

**Mohammad Nasir Uddin** Mohammad Nasir Uddin is a Visual Data Analyst and Applied AI Researcher at Taskimpetus Inc., and an MBA candidate in Data Analytics at Westcliff University, Irvine, CA, USA. His research interests include explainable artificial intelligence, financial fraud detection, and development econometrics. ORCID: 0009-0009-0990-4616.

**Md Munna Aziz** Md Munna Aziz is a Data Analyst and Applied AI Researcher at Westcliff University, Irvine, CA, USA. His research interests include explainable artificial intelligence, financial analytics, and machine learning. ORCID: 0009-0008-4845-8340.